\documentclass{article}

\PassOptionsToPackage{numbers,sort&compress}{natbib}
\usepackage[preprint]{neurips_2025}

\usepackage[utf8]{inputenc} %
\usepackage{hyperref}       %
\usepackage{url}            %
\usepackage{booktabs}       %
\usepackage{amsfonts}       %
\usepackage{nicefrac}       %
\usepackage{microtype}      %
\usepackage{xcolor}         %

\usepackage{transparent} %
\usepackage{import}      %

\usepackage{enotez}
\setenotez{
  backref=true,          %
  list-name={End Notes}  %
}

\usepackage{amsmath,amssymb,amsthm}
\usepackage{graphicx}
\usepackage{cleveref}
\usepackage[ruled,vlined]{algorithm2e}
\usepackage{bm}
\usepackage{enumitem}

\newcommand{\R}{\mathbb{R}}
\newcommand{\E}{\mathbb{E}}

\newcommand{\coloneqq}{\mathrel{\mathop:}=}

\usepackage[most]{tcolorbox}
\definecolor{keybg}{RGB}{252,248,240} %

\newtcolorbox{keybox}{
  colback=keybg,
  colframe=black,
  boxrule=1pt,                 %
  arc=2mm,
  left=6pt,right=6pt,top=6pt,bottom=6pt,
  enhanced,breakable,
  width=0.75\linewidth,         %
  before=\par\smallskip\centering,  %
  after=\par\smallskip
}

\title{Understanding sparse autoencoder scaling in the presence of feature manifolds}

\author{%
  Eric J. Michaud\textsuperscript{\dag,\ddag}\thanks{Correspondence: \texttt{ericjm@mit.edu}}\quad
  Liv Gorton\textsuperscript{\dag}\quad
  Tom McGrath\textsuperscript{\dag} \\
  \textsuperscript{\dag}Goodfire \qquad
  \textsuperscript{\ddag}Massachusetts Institute of Technology
}

\begin{document}

\maketitle

\begin{abstract}

Sparse autoencoders (SAEs) model the activations of a neural network as linear combinations of sparsely occurring directions of variation (latents). 
The ability of SAEs to reconstruct activations follows scaling laws w.r.t.\ the number of latents. In this work, we adapt a capacity-allocation model from the neural scaling literature (Brill, 2024) to understand SAE scaling, and in particular, to understand how \emph{feature manifolds} (multi-dimensional features) influence scaling behavior.
Consistent with prior work, the model recovers distinct scaling regimes. Notably, in one regime, feature manifolds have the pathological effect of causing SAEs to learn far fewer features in data than there are latents in the SAE. We provide some preliminary discussion on whether or not SAEs are in this pathological regime in the wild.

\end{abstract}

\section{Introduction}

Sparse autoencoders~\cite{ng2011sparse,cunningham2023sparse,bricken2023monosemanticity,gao2024scaling,rajamanoharan2024jumping,bussmann2024batchtopk,fel2025archetypal,muchane2025incorporating,costa2025flat} and related methods~\cite{dunefsky2024transcoders,lindsey2024sparse,gorton2024group, minder2025robustly, ameisen2025circuit} decompose neural network activations into a collection of sparsely activating latents. As SAEs have been scaled (now to millions of latents)~\cite{templeton2024scaling, gao2024scaling, lieberum2024gemma, lindsey2025biology}, they have exhibited \emph{scaling laws}~\cite{templeton2024scaling,lindsey2024scaling,gao2024scaling}, where loss improves predictably as a power law with the number of latents in the SAE. Although sparse autoencoders learn many interpretable features, they may miss important structure in neural representations, either because there are an extraordinary number of very rare features in activations~\cite{olah2024darkmatter} or because the SAE architecture and training objective make incorrect assumptions about the structure of neural representations~\cite{csordas2024recurrent,hindupur2025projecting}.

In this work, we develop a formal analysis of SAE scaling behavior and scaling laws. We are particularly interested in understanding SAE scaling when activations contain a particular kind of structure: \emph{feature manifolds} (multi-dimensional features)~\cite{engels2024not,modell2025origins}, and in whether feature manifolds could cause pathological scaling and exacerbate the problem of interpretability ``dark matter''~\cite{olah2024darkmatter,engels2024decomposing}. Our main approach is to adapt a mathematical model of \emph{neural scaling} from Brill (2024)~\cite{brill2024neural}, where models allocate capacity between different data manifolds, to the case of SAEs. Guided by our model, we then conduct experiments to probe whether SAEs may scale pathologically in practice.

\section{A model of sparse autoencoder scaling}\label{sec:model}

\subsection{The structure of data and the SAE architecture}\label{sec:model:data-and-architecture}

\textbf{Activation data}: We assume the multi-dimensional linear representation hypothesis~\cite{elhage2022superposition, engels2024decomposing} and that neural network activation vectors $\mathbf{x} \in \mathbb{R}^d$ are generated as a sum of sparsely occurring \emph{features}: $\mathbf{x} = \sum_{i} \mathbf{S}_i \mathbf{f}_i$ where $\mathbf{S}_i$ is the subspace where feature $i$ lives, specified by a $d \times d_i$ matrix whose columns are basis vectors of the subspace, and $\mathbf{f}_i$ is a random variable taking values in $\mathbb{R}^{d_i}$, where $d_i$ is the \emph{dimension} of feature $i$. Each feature $\mathbf{f}_i$ is \emph{sparse}, supported on a small fraction of the data: $p_i = \Pr[ \mathbf{f}_i \neq \mathbf{0} ] \ll 1$, so each activation vector is a sum of only a small subset of all features. 

\textbf{SAEs}: Sparse autoencoders attempt to reconstruct activation vectors as a sum of sparsely activating latents, consisting of an encoding step $\hat{\mathbf{f}} = {\rm Enc}(\mathbf{x}) = \sigma(\mathbf{W}_e \mathbf{x} + \mathbf{b}_e)$ where $\sigma$ is a nonlinearity and $\hat{\mathbf{f}} \in \R^N$, and a decoding step $\hat{\mathbf{x}} = {\rm Dec}(\hat{\mathbf{f}}) = \mathbf{W}_d \hat{\mathbf{f}} + \mathbf{b}_d$. Sparse autoencoders are trained with SGD on a loss $\mathcal{L} = \| \mathbf{x} - \hat{\mathbf{x}} \|_2^2 + \lambda\, S(\hat{\mathbf{f}})$ where $S(\hat{\mathbf{f}})$ is a sparsity-encouraging loss like $\| \hat{\mathbf{f}} \|_1$ or $\| \hat{\mathbf{f}} \|_0$.

\subsection{Assumption: SAE optimization reduces to a latent allocation problem}\label{sec:model:math}

\begin{figure}[t!]
    \centering
    \includegraphics[width=0.8\linewidth]{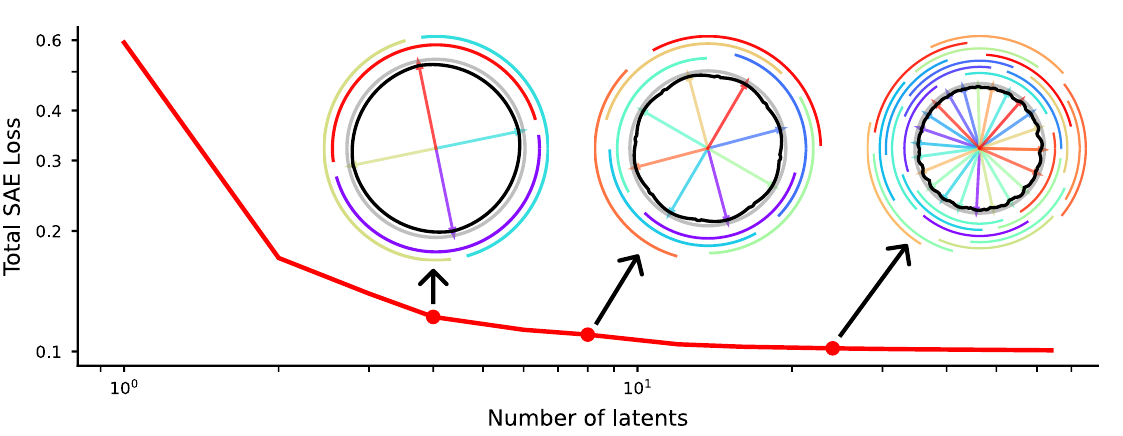}
    \caption{\textbf{SAE scaling on a toy feature manifold $S^1$}. We train ReLU SAEs with an L1 sparsity penalty to reconstruct points on the circle $S^1 \subseteq \R^2$. We find that SAEs can slightly reduce their total loss by ``tiling'' the manifold with more sparsely activating latents. For SAEs with 4, 8, and 24 latents, we show the data ($S^1$) in grey, the SAE's reconstruction in black, the decoder latent directions as arrows, and indicate the part of the circle that each latent fires on as colored arcs. If SAEs can reduce loss by ``tiling'' a manifold in this way, they may do this at the expense of learning rarer features.}
    \label{fig:manifolds-spoke-diagrams}
\end{figure}

We assume that SAEs learn solutions where each latent $j$ is \emph{specific} to a particular feature $i$, so that $\hat{\mathbf{f}}_j \neq 0$ only if $\mathbf{f}_i \neq \mathbf{0}$. We further assume that the SAE latent decoder directions for the latents associated with feature $i$ lie within the subspace $\mathbf{S}_i$, and that the subspaces where features live are orthogonal, i.e., $\mathbf{S}_i \perp \mathbf{S}_k$ if $i \neq k$. It follows that the SAE loss on any sample $\mathbf{x}$ is the sum of losses on samples where each active feature had fired alone: if $\mathcal{A}(\mathbf{x})$ are the ``active'' features $i$ where $\mathbf{f}_i \neq \mathbf{0}$ on $\mathbf{x}$, then
$$ \mathcal{L}(\textbf{x}) = \mathcal{L}\left( \sum_{i \in \mathcal{A}(\mathbf{x})} \mathbf{S}_i \mathbf{f}_i\right) \overset{\text{assumption}}{=} \sum_{i\in \mathcal{A}(\mathbf{x})} \mathcal{L}(\mathbf{S}_i \mathbf{f}_i).$$
We note that feature absorption and related phenomena~\cite{chanin2024absorption} and the non-orthogonality of features in practice violate these assumptions, but we will accept them for the sake of expedience. If SAE latents are specific to features, then an optimal SAE's loss on a feature $i$ will be determined by the number of latents $n_i$ the SAE allocates to reconstructing feature $i$, which we denote $L_i(n_i)$, and will be determined by the geometry of the feature $\mathbf{f}_i$.
The expected loss across activations $\mathbf{x}$ is then:
$$ \mathbf{E}_\mathbf{x}\left[ \mathcal{L}(\mathbf{x}) \right] = \mathbf{E}_\mathbf{x}\left[  \sum_{i\in \mathcal{A}(\mathbf{x})} \mathcal{L}(\mathbf{S}_i \mathbf{f}_i) \right] = \sum_i p_i \mathcal{L}(\mathbf{S}_i \mathbf{f}_i) = \boxed{\sum_i p_i L_i(n_i)} $$

We therefore reduce the SAE optimization problem to the problem of choosing how many latents $n_i$ to allocate to each feature $i$ in the data. To understand SAE scaling behavior, we solve for the allocation of latents to each feature $n_i$ that minimizes $\sum_i p_i L_i(n_i)$ given a distribution over feature frequencies $p_i$, the per-feature loss curves $L_i$, and the constraint on the total number of latents $\sum_i n_i = N$.

\begin{figure}
    \centering
    \includegraphics[width=\linewidth]{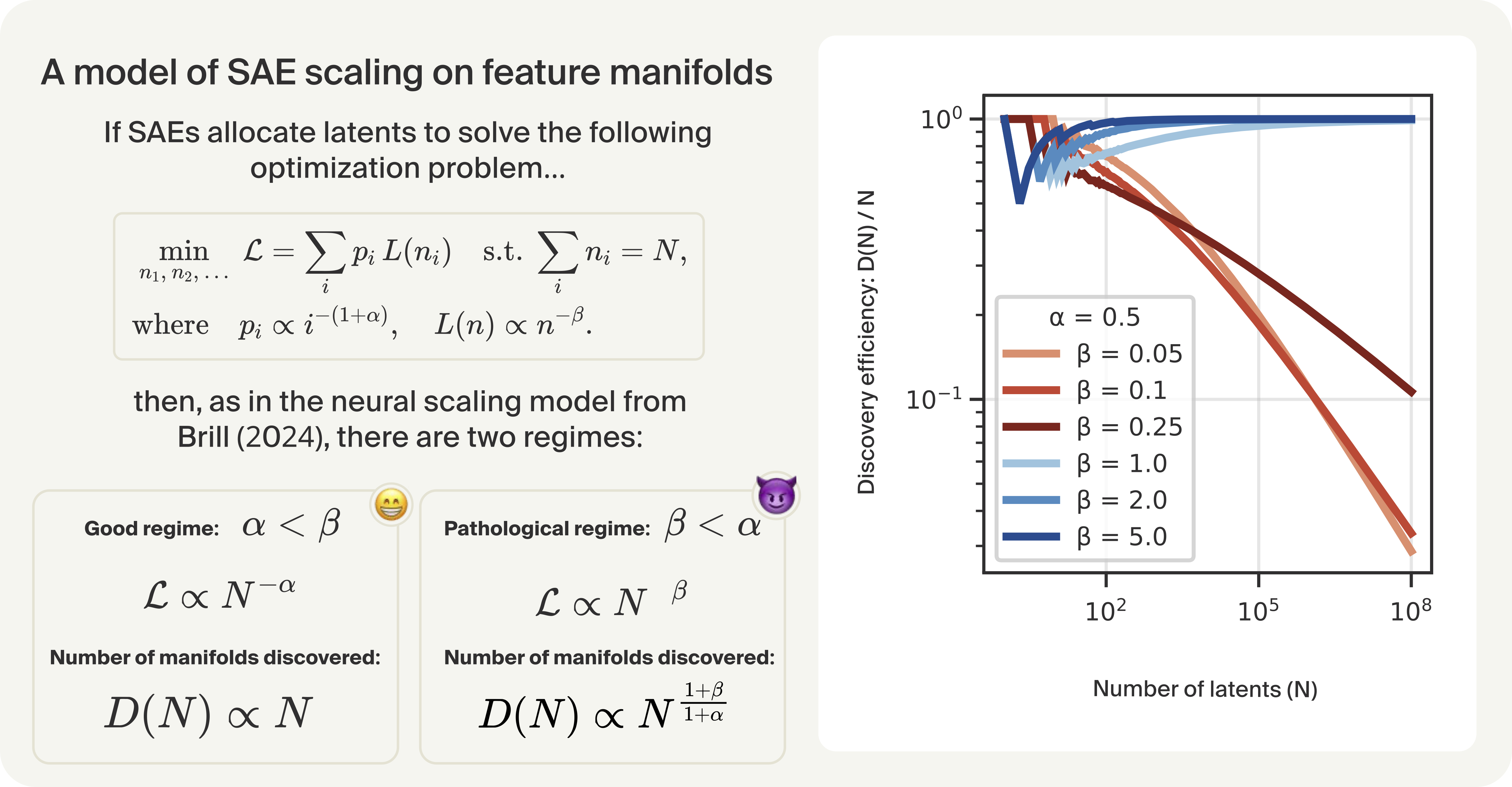}
    \caption{\textbf{Left}: Application of Brill's (2024)~\cite{brill2024neural} capacity-allocation model to SAE scaling. \textbf{Right}: numerical simulation of SAE scaling when the most frequently occurring feature is a manifold with loss scaling as $L(n) \propto n^{-\beta}$ and all other features are discrete. We see that if $\beta \ll \alpha$, then a simulated SAE with 100 million latents discovers only 3 million features.}
    \label{fig:graphic-sae-scaling-fig}
\end{figure}

\subsection{Warm-up on discrete features}\label{sec:model:warm-up}

To get comfortable with this formulation, we first apply it to the simplest case where all features are \emph{discrete}. We imagine that for all features $i$, $d_i = 1$ and $\mathbf{f}_i = 1$ on the fraction $p_i$ of samples where feature $i$ is present and is $0$ otherwise. For each feature $i$, an SAE incurs loss $1$ if $n_i = 0$ (reconstruction loss = 1 and sparsity loss = 0). If $n_i = 1$, it can perfectly reconstruct the feature, incurring sparsity loss $\lambda$ with L0 or L1 sparsity loss. Therefore $L(n) = 1 \ {\rm if }  \ n_i < 1 \ {\rm else} \ \lambda$.

With total loss $\sum_i p_i L(n_i)$ and $N$ latents in our SAE we see that the optimal solution is to allocate one latent to the most commonly occurring $N$ features. In this setup, then, \textbf{the effect of scaling SAEs is to learn an increasing number of discrete features in the data, in decreasing order of their frequency}. We will often be interested in the number of features ``discovered'' by the SAE $D(N) = | \{ i \ : \ n_i > 0 \} |$. In this setting, $D(N) = N$. To recover the power law scaling that SAEs exhibit empirically, we only need to assume that $p_i \propto i^{-(1 + \alpha)}$. With this assumption, the improvement in total loss from adding a marginal latent $i$ follows $p_i (1 - \lambda)$, and integrating from $i = 1\ldots N$ we get that the total loss drops off as a power law $\mathcal{L}(N) \propto N^{-\alpha}$. We note that this model of SAE scaling mirrors the ``quanta'' model of neural scaling from~\cite{michaud2023quantization}, where here the ``quanta'' are features. This picture also agrees with the finding from~\cite{templeton2024scaling} that SAEs learn features for concepts in data approximately in order of the frequencies (roughly Zipfian) at which those concepts occur.

\subsection{Intuition behind pathological manifold scaling}\label{sec:model:intuition}

Recently, several works have commented on the existence of \emph{feature manifolds} ~\cite{modell2025origins, gorton2024group} (multi-dimensional features)~\cite{engels2024not} in neural networks. In our formalism above, these are features $i$ with $d_i > 1$ and where the range of $\mathbf{f}_i$ is a manifold embedded in $\R^{d_i}$. With feature manifolds, instead of having a discrete $L_i(n_i)$ curve like above, $L_i$ might drop off slowly as $n_i$ grows. To show that this is possible, in~\Cref{fig:manifolds-spoke-diagrams}, we show the scaling curve for ReLU SAEs ($\sigma = {\rm ReLU}$) trained with L1 penalty to reconstruct points on a circle $\mathbf{x} \in S^1 \subseteq \R^2$. We observe that these SAEs can gradually reduce their total loss by more finely ``tiling'' the manifold with latents that activate more sparsely, and that this manifold can accommodate dozens of latents before loss plateaus. While such solutions ruthlessly minimize the SAE loss, it is not obvious that they would be better from an interpretability standpoint. Our fundamental concern is that if $L_i(n_i)$ curves decrease gradually, it could be optimal for an SAE to tile common feature manifolds instead of discovering rarer features in data.

\begin{figure}
    \centering
    \includegraphics{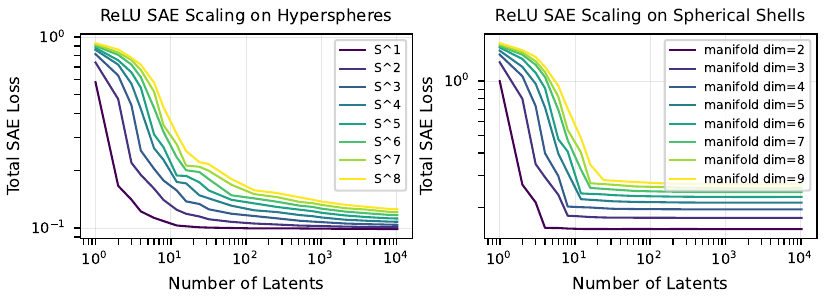}
    \caption{ReLU SAE scaling on individual toy feature manifolds, showing how $L(n)$ curves depend on feature geometry. We train with L1 sparsity $\lambda = 0.1$. \textbf{Left}: SAE scaling on data drawn from a single unit hypersphere. \textbf{Right}: SAE scaling on data sampled in $\{ \mathbf{x} \ : \ 0.5 < |\mathbf{x}| < 2 \} $.}
    \label{fig:relu-scaling-toymanifolds-both}
\end{figure}

\subsection{Solution for power-law L(n) following Brill (2024)}\label{sec:model:results}

To model SAE scaling in the presence of feature manifolds, we assume that feature frequencies decay as a power law $p_i \propto i^{-(1 + \alpha)}$. For mathematical simplicity, we also assume that all features have the same power law per-feature loss curve $L_i(n_i) = n_i^{-\beta}$. 
In this setting, our model of SAE scaling directly corresponds to the model of neural scaling from Brill (2024)~\cite{brill2024neural}, where neural networks allocate units of capacity $n_i$ towards approximating functions on distinct power-law distributed data manifolds, with per-manifold loss scaling as $n_i^{-c/D}$.
We show similar derivations to Brill~\cite{brill2024neural} in~\Cref{sec:appendix:derivations} and summarize the core results in~\Cref{fig:graphic-sae-scaling-fig} (\textbf{left}). 

In our notation, there is a key threshold: $\alpha < \beta$. When $\alpha < \beta$, 
expected loss scales as $\mathcal{L}(N) \propto N^{-\alpha}$ like in the discrete feature case, and 
$D(N) \propto N$, so the ``efficiency'' at which SAEs discover features $D(N) / N$ tends towards a constant. However, when $\beta < \alpha$, then 
$\mathcal{L}(N) \propto N^{-\beta}$ and 
$D(N) \propto N^{\frac{1 + \beta}{1 + \alpha}}$, and so $D(N) / N$ tends towards 0. This illustrates the core dynamic of concern: when SAEs can continue to reduce loss by ``tiling'' common feature manifolds, then it can be optimal for them to do this at the expense of discovering other, rarer features.

\subsection{Numerical simulation}\label{sec:model:numerics}

While our mathematical model above assumed that all features have the same $L_i(n_i)$ curve, we can run numerical simulations with arbitrary heterogeneous $L_i(n_i)$ curves. In~\Cref{fig:graphic-sae-scaling-fig} (\textbf{right}), we conduct simulations where the first feature scales as $n^{-\beta}$ but all other features are discrete and scale as step functions $\mathbf{1}_{n = 0}$, and with $p_i \propto i^{-(1 + 0.5)}$. We find that, with only a single feature with power-law $L(n_i)$ scaling, when $\beta < \alpha$ this feature begins to absorb the vast majority of latents in the SAE once $N$ is large.

\section{SAE scaling on synthetic features and on real neural networks}\label{sec:sae-hyperspheres}

In our analysis, whether SAEs will scale pathologically depends on the shape of their per-feature loss curves $L_i(n_i)$.
We showed one such $L(n)$ curve in~\Cref{fig:manifolds-spoke-diagrams}, but we further explore this by training SAEs on a variety of synthetic manifolds in~\Cref{fig:relu-scaling-toymanifolds-both} and in Appendix~\Cref{fig:jumprelu-scaling-toymanifolds-both}. Overall, we find that the shape of the $L(n)$ curve depends on the manifold geometry, with some manifolds accommodating many thousands of latents without saturating loss while on others the loss curve plateaus very quickly (effectively $\beta \rightarrow \infty$). In~\Cref{sec:appendix:additional-discussion:pathological}, we provide further discussion on what $\alpha$ and $\beta$ may be, and whether SAEs might be scaling pathologically, in practice.

When a large number of SAE latents are allocated to a relatively low-dimensional feature manifold, we'd expect the cosine similarities between the decoder latents allocated to that manifold to be close to $1$. In~\Cref{sec:appendix:llm-and-vision-sae-geometry}, we show the distribution over decoder latent nearest neighbor cosine similarities for SAEs trained on LLM and vision model activations, and find some differences between them.

\section{Discussion}\label{sec:discussion}

In this short paper, we have described a model of SAE scaling which reduces the SAE optimization problem to the problem of optimally allocating different numbers of SAE latents to different features in data. As in the model of scaling from Brill~\cite{brill2024neural} SAE scaling laws either result from an underlying power law distribution over features $p_i \propto i^{-(1 + \alpha)}$, or from the improvements in loss from tiling common feature manifolds following a power law $L_i(n_i) \propto n_i^{-\beta}$. When $\beta < \alpha$, SAE latents could massively accumulate on commonly occurring feature manifolds.

Unfortunately, we do not resolve the question of whether SAEs are in this pathological scaling regime in practice. Our uncertainty is due to our knowing neither the distribution over true feature frequencies (determining $\alpha$) nor the geometry of real-world neural network feature manifolds (which determines $\beta$). We give some additional commentary in~\Cref{sec:appendix:additional-discussion:pathological}. Arguably our most important observation is that when we train SAEs on toy feature manifolds with variation in the radial direction, our SAEs do not tile the manifold and instead exhibit $L(n)$ curves which plateau early. If this geometry is realistic~\cite{engels2024not,olah2024what}, then manifolds may not pose an issue to SAE scaling in practice. Overall though, we view this work as primarily about framing, rather than completely answering, this interesting problem.

\begin{ack}

We thank Owen Lewis, Michael Pearce, Jack Merullo, Dan Balsam, Michael Byun, Elana Simon, and Wes Gurnee for helpful conversations and feedback. EJM was supported by the NSF via the Graduate Research Fellowship Program (Grant No.2141064). This work was done at \href{https://goodfire.ai}{Goodfire AI} in San Francisco.

\end{ack}

\bibliographystyle{unsrtnat}
\bibliography{refs}

\appendix

\section{Additional discussion}\label{sec:appendix:additional-discussion}

\subsection{Are real SAEs in the pathological regime?} \label{sec:appendix:additional-discussion:pathological}

It is worth attempting to say more about whether SAEs are in the pathological scaling regime in practice. As we stated in~\Cref{sec:model:results}, this depends on the rate at which the feature frequencies $p_i \propto i^{-(1 + \alpha)}$ decay vs. the rate at which the per-feature SAE loss decays $L(n_i) \propto i^{-\beta}$. In~\Cref{sec:appendix:derivations}, following Brill (2024)~\cite{brill2024neural}, we show that when $\alpha < \beta$, the efficiency at which SAEs discover features $D(N) / N$ approaches a reasonable constant, but when $\beta < \alpha$, $D(N) / N$ approaches 0 as $N \rightarrow \infty$. Whether feature manifolds could cause pathological SAE scaling in the real world depends then on the real $\alpha$ (assuming it's even a power law) and $\beta$ (assuming $L_i(n_i)$ is also a power law $\propto n_i^{-\beta}$). 

\begin{figure}[h]
    \centering
    \includegraphics{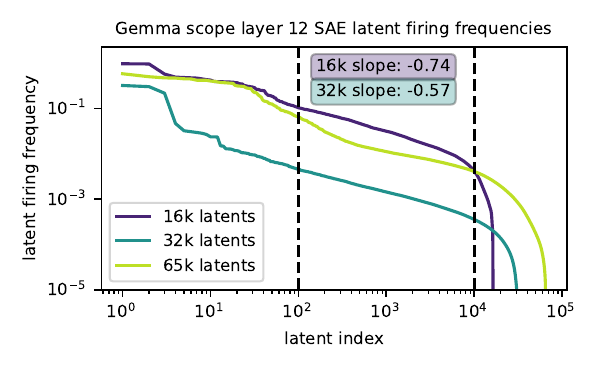}
    \caption{Frequencies at which latents fire in gemma scope SAEs, sorted by frequency. We measure the power law decay exponent between latent $10^2$ and $10^4$ to be -0.74 for an SAE with 16k latents and -0.57 for an SAE with 32k latents.}
    \label{fig:gemma-scope-frequencies}
\end{figure}

\textbf{What is $\alpha$?}: We first speculate on what $\alpha$ may be. One way of trying to measure this is to look at how the latent activation frequencies decay when sorted by frequency. For a few Gemma Scope SAEs~\cite{lieberum2024gemma}, we show these curves in~\Cref{fig:gemma-scope-frequencies}, and measure slopes between $-0.57$ and $-0.74$. If there was a one-to-one relationship between SAE latents and features in the data, then this would imply an $\alpha \approx 0.5$ to $\alpha \approx 0.7$. However, feature absorption~\cite{chanin2024absorption}, the learning of compositional features~\cite{till2024true, anders2024sparse}, and latents tiling a feature manifold could distort the relationship between the true feature frequencies and the latent activation frequencies. 

Another highly speculative way of trying to estimate $\alpha$ could be to look at the exponents of neural scaling laws for models like those the SAE is being trained on. The idea here is that if the ``features'' are the computational units of neural networks that ref~\cite{michaud2023quantization} called the ``quanta'', then the underlying neural scaling law slope would reflect the distribution over feature occurrences. Neural scaling exponents for language models (w.r.t. network parameters) ($\alpha_N$ in the scaling law $N^{-\alpha_N}$ have been measured to have an $\alpha_N$ of $0.07$~\cite{kaplan2020scaling} and $0.34$~\cite{hoffmann2022training}, potentially implying a distribution over quanta/features $p_i \propto i^{-(1 + \alpha)}$ with exponent $\alpha$ in that same range.

\begin{figure}
    \centering
    \includegraphics[width=0.9\textwidth]{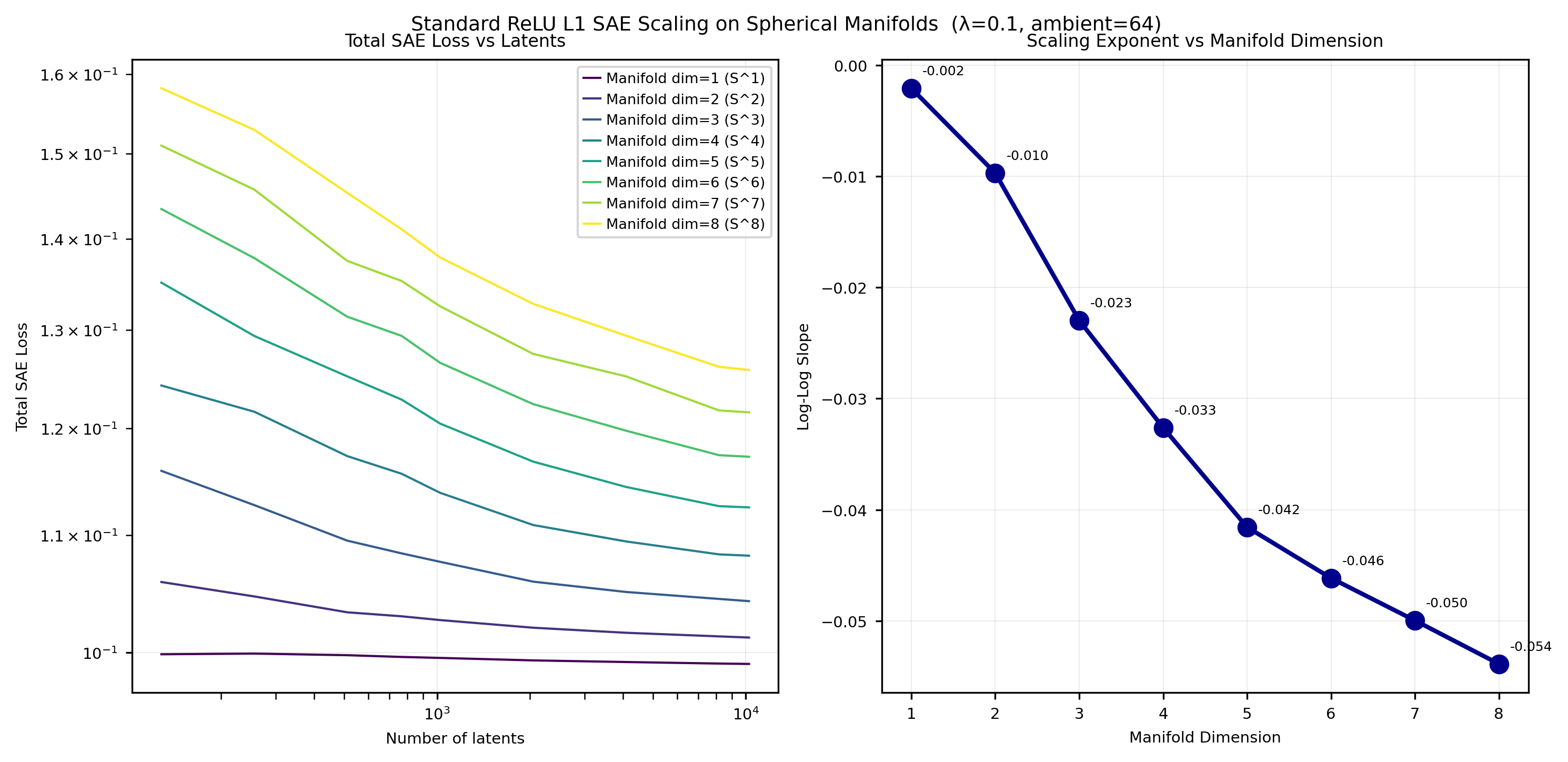}
    \caption{Measured slopes of $L_i$ curves for ReLU L1 SAEs trained on hyperspheres.}
    \label{fig:relu-spherical-manifold-slopes}
\end{figure}

\textbf{What is $\beta$?}: In~\Cref{fig:relu-scaling-toymanifolds-both} and~\Cref{fig:jumprelu-scaling-toymanifolds-both}, we show $L(n)$ scaling curves for SAEs reconstructing single synthetic feature manifolds. We find that these curves depend on the feature geometry. In particular, we see that when reconstructing hollow hyperspheres, we can observe gradual $L(n)$ scaling. The higher-dimensional hyperspheres in particular can accommodate at least $10^4$ latents without loss plateauing. In~\Cref{fig:relu-spherical-manifold-slopes}, we plot the slope that we measure for these curves between $10^2$ and $10^4$ latents, and measure a $\beta$ of roughly $0.05$ for hyperspheres with dimension $6$-$8$.

However, in the more realistic setting where there is variation in the radial direction--the intensity that features fire~\cite{olah2024what}--we see that manifolds tend to saturate very quickly. It appears that the saturation happens when $n_i \approx 2d_i$, likely corresponding to solutions where the SAE latents form a basis for the subspace where the feature is embedded (or rather use two latents for each basis direction, one in the ``positive'' direction and one in the ``negative'' direction since latents can only fire positively). 

Therefore, the slope of the $L_i(n_i)$ curve, and whether SAEs can use a large number of latents to reduce loss on feature manifolds, depends on the geometry of the manifold. Our experiments so far indicate that in the more realistic setting where there is variation in the radial direction (which was seen in practice in ref~\cite{engels2024not}), that SAEs do not discover solutions which take advantage of a large number of latents, and instead learn a basis solution. \textbf{This is probably the strongest argument against the possibility of feature manifolds causing pathological SAE scaling.}

Lastly, we note that we are unsure how ``ripples''~\cite{olah2024what} in feature manifolds could affect SAE scaling on them. If a feature manifold is intrinsically low dimensional, but ripples through a large number of other dimensions, we could imagine SAE solutions potentially looking different.

\section{Derivations}\label{sec:appendix:derivations}

\subsection{Loss decomposition into per-feature terms}
We work under the assumptions in \S\ref{sec:model:math}: (i) \emph{feature-specific latents}—each latent $j$ fires only when a unique feature $i(j)$ is active; (ii) \emph{decoder respect for subspaces}—decoder columns for latents assigned to feature $i$ lie in $\mathrm{span}(\mathbf{S}_i)$; and (iii) \emph{orthogonal feature subspaces}—$\mathrm{span}(\mathbf{S}_i)\perp \mathrm{span}(\mathbf{S}_k)$ for $i\neq k$.

\paragraph{Additivity of sparsity.}
For a sample $\mathbf{x}=\sum_{i\in\mathcal{A}(\mathbf{x})}\mathbf{S}_i\mathbf{f}_i$, feature-specificity implies $\hat{\mathbf{f}}_j(\mathbf{x})=0$ unless $i(j)\in\mathcal{A}(\mathbf{x})$. For separable sparsity penalties (L0/L1), $S(\hat{\mathbf{f}})=\sum_j s(\hat{f}_j)$, so
\[
S(\hat{\mathbf{f}}(\mathbf{x}))=\sum_{i\in\mathcal{A}(\mathbf{x})}\;\sum_{j:\,i(j)=i}s(\hat{f}_j(\mathbf{x})) \, .
\]
Thus the sparsity cost splits across active features.

\paragraph{Orthogonal reconstruction.}
Write the model reconstruction as $\hat{\mathbf{x}}=\sum_i \hat{\mathbf{x}}_i$, with
$\hat{\mathbf{x}}_i\coloneqq \sum_{j:\,i(j)=i}\mathbf{w}_j\hat{f}_j\in \mathrm{span}(\mathbf{S}_i)$ by (ii).
Then, using (iii) and the Pythagorean theorem,
\[
\|\mathbf{x}-\hat{\mathbf{x}}\|_2^2
=\Big\|\sum_{i}\big(\mathbf{S}_i\mathbf{f}_i-\hat{\mathbf{x}}_i\big)\Big\|_2^2
=\sum_{i}\|\mathbf{S}_i\mathbf{f}_i-\hat{\mathbf{x}}_i\|_2^2 \, .
\]
Hence the per-sample objective
$\mathcal{L}(\mathbf{x})=\|\mathbf{x}-\hat{\mathbf{x}}\|_2^2+\lambda S(\hat{\mathbf{f}})$
decomposes as a sum over features active on that sample. Taking expectations and letting
$n_i$ be the number of latents allocated to feature $i$, we obtain
\begin{equation}
\boxed{\quad \E_{\mathbf{x}}\big[\mathcal{L}(\mathbf{x})\big]
= \sum_i p_i\, L_i(n_i), \qquad \sum_i n_i=N \quad}
\label{eq:master-objective}
\end{equation}
where $p_i\coloneqq \Pr[\mathbf{f}_i\neq 0]$ and $L_i(n_i)$ is the optimal (feature-$i$) expected reconstruction-plus-sparsity loss achieved with $n_i$ latents restricted to $\mathrm{span}(\mathbf{S}_i)$. We note that the derivations below closely follow those in Brill (2024)~\cite{brill2024neural}, adapted to our SAE setting with appropriate changes in notation and interpretation. We show them here for convenience.

\subsection{Scaling setup and notation}
We study the optimal latent allocation $n_i$ minimizing \eqref{eq:master-objective} under two empirical power-law regularities:
\[
p_i \propto i^{-(1+\alpha)} \quad\text{(features sorted by frequency)},\qquad
L_i(n)\equiv L(n)\propto n^{-\beta} \, ,
\]
with $\alpha,\beta>0$.\footnote{Constants are inessential for power-law exponents and are dropped.}
Define the discovery count
\[
D(N)\coloneqq \big|\{\,i:\;n_i>0\,\}\big|
\]
and the total expected loss $\mathcal{L}(N)\coloneqq \sum_i p_i L(n_i)$ at total width $N$.
A standard Lagrange multiplier treatment (continuous relaxation) yields
\begin{equation}
n_i \;\propto\; p_i^{\frac{1}{1+\beta}} \;\propto\; i^{-\gamma}, 
\qquad \gamma \coloneqq \frac{1+\alpha}{1+\beta}.
\label{eq:allocation}
\end{equation}
In practice there is a cutoff index $i_c$ (``last discovered feature'') with $n_{i_c}\approx 1$ and $n_i\lesssim 1$ for $i>i_c$. Then $D(N)\asymp i_c$ and
\[
N=\sum_{i\le i_c} n_i \;\propto\; \sum_{i\le i_c} i^{-\gamma}.
\]
Two regimes follow depending on whether the allocation tail-sum diverges or converges.

\subsection{Case \texorpdfstring{$\beta<\alpha$}{beta<alpha} (simple, latent accumulation on frequent features)}

Here $\gamma>1$, so $\sum_{i\ge 1} i^{-\gamma}$ converges to a constant $Z(\gamma)$. From \eqref{eq:allocation},
$N\propto \sum_{i\le i_c} i^{-\gamma}\to Z(\gamma)$ implies the proportionality constant in \eqref{eq:allocation} scales as $\kappa\propto N$. The discovery cutoff is set by $n_{i_c}\approx 1$:
\[
1 \;\approx\; \kappa\, i_c^{-\gamma}
\quad\Longrightarrow\quad
i_c \;\propto\; \kappa^{1/\gamma} \;\propto\; N^{\frac{1}{\gamma}}
= N^{\frac{1+\beta}{1+\alpha}}.
\]
Thus
\begin{equation}
\boxed{\quad D(N)\;\propto\; N^{\frac{1+\beta}{1+\alpha}} \quad (\text{sublinear discovery}). \quad}
\end{equation}
For the loss, the discovered part scales as
\[
\sum_{i\le i_c} p_i\, n_i^{-\beta} \;\propto\; \kappa^{-\beta}\!\sum_{i\le i_c} i^{-\gamma}
\;\propto\; \kappa^{-\beta}\;\propto\; N^{-\beta},
\]
while the undiscovered tail $\sum_{i>i_c} p_i \propto i_c^{-\alpha}\propto N^{-\alpha(1+\beta)/(1+\alpha)}$ decays faster since $\alpha>\beta$. Therefore
\begin{equation}
\boxed{\quad \mathcal{L}(N)\;\propto\; N^{-\beta}. \quad}
\end{equation}
Intuitively, the SAE keeps shaving loss on common feature manifolds; discovery lags.

\subsection{Case \texorpdfstring{$\alpha<\beta$}{alpha<beta} (benign, feature discovery keeps up)}
Now $\gamma<1$, so $\sum_{i\le i_c} i^{-\gamma}\propto i_c^{1-\gamma}$. Using $N\propto \kappa\, i_c^{1-\gamma}$ and the threshold $1\approx \kappa\, i_c^{-\gamma}$, we eliminate $\kappa$ to find
\[
N \;\propto\; i_c \qquad\Longrightarrow\qquad \boxed{\; D(N)\;\propto\; N. \;}
\]
For the loss over discovered features,
\[
\sum_{i\le i_c} p_i\, n_i^{-\beta}
\;\propto\; \kappa^{-\beta} \sum_{i\le i_c} i^{-\gamma}
\;\propto\; i_c^{-\beta\gamma}\, i_c^{1-\gamma}
\;=\; i_c^{\,1-(1+\alpha)} \;=\; i_c^{-\alpha}
\;\propto\; N^{-\alpha}.
\]
The undiscovered tail obeys $\sum_{i>i_c} p_i \propto i_c^{-\alpha}\propto N^{-\alpha}$, so both pieces match and
\begin{equation}
\boxed{\quad \mathcal{L}(N)\;\propto\; N^{-\alpha}. \quad}
\end{equation}
Here, extra width primarily buys new features rather than over-tiling old manifolds; loss scaling mirrors the frequency tail.

\section{Additional Experimental Details}

For~\Cref{fig:relu-scaling-toymanifolds-both} and~\Cref{fig:jumprelu-scaling-toymanifolds-both}, we trained SAEs on synthetic feature manifolds. For these experiments, we trained for 12000 steps, with a batch size of 2048, and a learning rate of $10^{-3}$ with the Adam optimizer. For the L1 penalty calculation, we use the trick of multiplying the decoder vector L2 norms by the latent activation~\cite{conerly2024saeupdate}.

\section{SAE scaling curves on synthetic manifolds}

\begin{figure}
    \centering
    \includegraphics{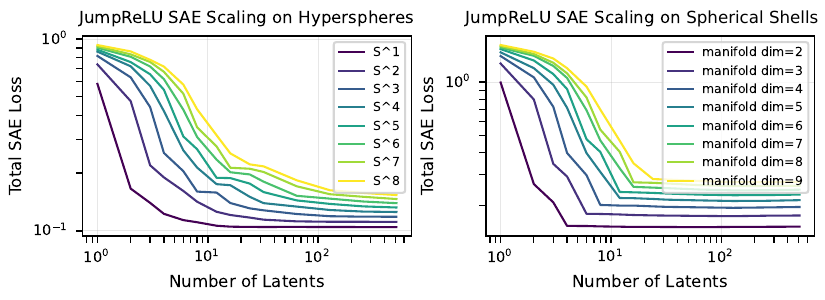}
    \caption{JumpReLU SAE scaling on individual toy feature manifolds, showing how $L(n)$ curves depend on feature geometry. We train with the tanh loss from~\cite{conerly2025dictionary} with $c = 0.1$ and $\lambda_s = 1.0$. \textbf{Left}: JumpReLU SAE scaling on unit hyperspheres of varying dimension. \textbf{Right}: JumpReLU SAE scaling on points sampled in $\{ \mathbf{x} \ : \ 0.5 < |\mathbf{x}| < 2 \} $. We see that, like with ReLU SAEs, that when there is variation in the radial direction between samples that our SAEs do not learn solutions which can continue to accommodate latents, and instead plateau after allocating roughly $2d_i$ latents to the feature manifold.}
    \label{fig:jumprelu-scaling-toymanifolds-both}
\end{figure}

In~\Cref{fig:jumprelu-scaling-toymanifolds-both}, we show JumpReLU SAE scaling on individual feature manifolds like we did for ReLU SAEs in~\Cref{fig:relu-scaling-toymanifolds-both}.

\section{SAE feature geometry on LLMs and vision models}\label{sec:appendix:llm-and-vision-sae-geometry}

\subsection{JumpReLU Gemma Scope SAEs}

If a large number of latents ``tile'' a single low-dimensional feature manifold, then we would expect the decoder directions for those SAE latents to have neighbors with high cosine similarity. One can see this effect directly in~\Cref{fig:manifolds-spoke-diagrams}, where we see that the SAE decoder latents begin to be arranged quite tightly together along the manifold. In this section, we study whether SAEs on real neural network activations have large numbers of latents with very high cosine similarity to their nearest neighbor. 

\begin{figure}[h!]
    \centering
    \includegraphics{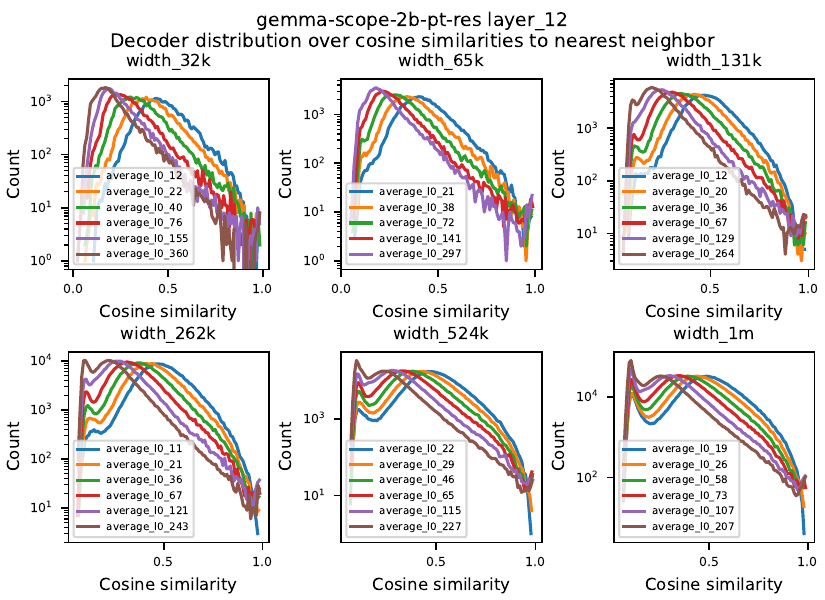}
    \caption{Distribution over cosine similarities between decoder vectors and their nearest neighbor.}
    \label{fig:decoder-cosine-similarities-gemma-scope-2b-pt-res-layer-12}
\end{figure}

We first study this in the Gemma Scope SAEs~\cite{lieberum2024gemma}. In~\Cref{fig:decoder-cosine-similarities-gemma-scope-2b-pt-res-layer-12}, we plot the distribution over cosine similarities between decoder latent vectors and their nearest neighbor for Gemma Scope SAEs on layer 12 (residual stream) of gemma-2-2b. While the distribution is skewed substantially higher than one would expect if all latent decoder vectors were random (and thus approximately orthogonal), we do not overall see a very large fraction of latents with extremely high cosine similarity to their nearest neighbor.

Intriguingly though, for some SAEs we do see a small uptick on the right side of this distribution, where between 10-100 latents have cosine similarity $> 0.97$ with their nearest neighbor. When we investigated these latents in one SAE (width\_262k, average\_l0\_121), we found that for each of these latents, their nearest neighbor was dead (across a dataset of over 250 million tokens). These latents are not then being used to very sparsely reconstruct points on a manifold, and instead seem to be an artifact of the training process. However, the alive latents in this set are not typical SAE latents. A large number of these latents fire on tokens representing single numerical digits and single alphabet characters. We do not have an explanation of this phenomenon, but wonder whether there may be some underlying manifold representation which the SAE at one point in training tried to ``tile'', but then when the latents got too close, one of them was killed to reduce the L0 loss. 

\subsection{ReLU L1 SAEs on Inception-v1}

We also study the geometry of latent decoder directions on SAEs trained on Inception-v1 activations. We train SAEs on activations from mixed3b using an L1 coefficient, $\lambda$, of 1, a learning rate of $10^{-4}$, and an expansion factor of 16 (a total of 7680 latents).

\begin{figure}[h!]
    \centering
    \includegraphics[width=0.7\linewidth]{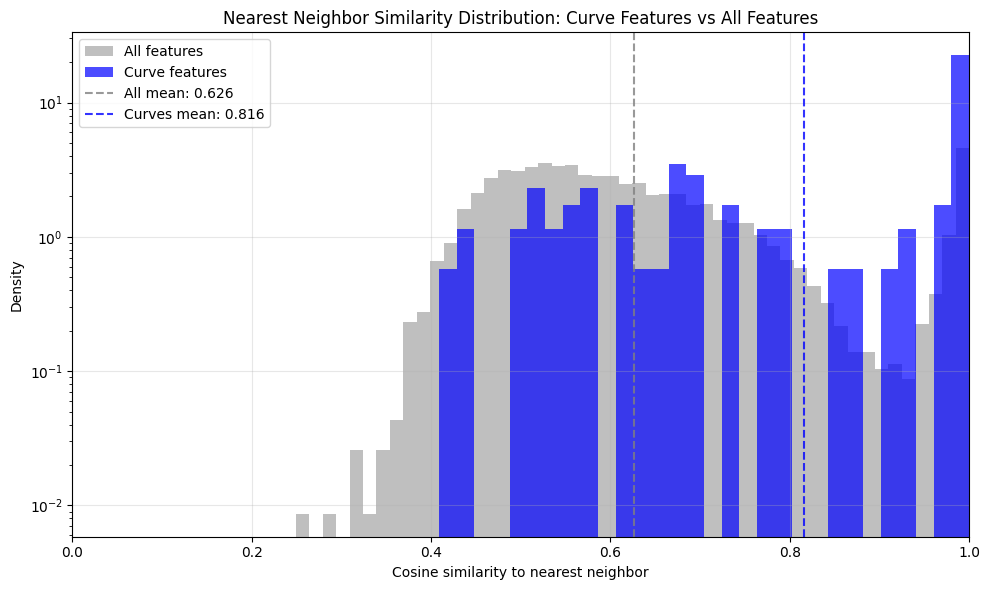}
    \caption{Distribution over pairwise cosine similarities for Inception V1.}
    \label{fig:liv-inception-cos-sim}
\end{figure}

In~\Cref{fig:liv-inception-cos-sim}, we find that on Inception-v1, a meaningful fraction of SAE latents have very high cosine similarity with their nearest neighbor. We note however that this could be due to latents being duplicated, which is not strongly disincentivized by the L1 loss, as pointed out in~\cite{minder2025robustly}.

\end{document}